%% file: main.tex
\documentclass[journal]{IEEEtran}
\usepackage{graphicx} 

\usepackage{caption} 
\usepackage{amssymb}
\usepackage{xcolor}
\usepackage{algorithm, algorithmic}
\usepackage[switch]{lineno}  %
\usepackage{amsmath}
\input{resources.tex}

\newcommand\xl[1]{\textcolor{black}{#1}}

\DeclareMathAlphabet\mathbfcal{OMS}{cmsy}{b}{n}

\begin{document}

\title{Self-Training for Class-Incremental \\ Semantic Segmentation}

\author{Lu Yu, Xialei Liu, and Joost Van de Weijer
\thanks{Lu Yu is with School of Computer Science and Engineering, Tianjin University of Technology, China, 300384, e-mail:luyu@email.tjut.edu.cn.}

\thanks{Xialei Liu is the corresponding author, with College of Computer Science, Nankai University, China,300071, e-mail:xialei@nankai.edu.cn.}

\thanks{Joost van de Weijer is with Computer Vision Center, Autonomous University of Barcelona, Spain, 08193, e-mail:joost@cvc.uab.es.}
}


\maketitle

\begin{abstract}
In class-incremental semantic segmentation we have no access to the labeled data of previous tasks. Therefore, when incrementally learning new classes, deep neural networks suffer from catastrophic forgetting of previously learned knowledge. To address this problem, we propose to apply a self-training approach that leverages unlabeled data, which is used for rehearsal of previous knowledge. Specifically, we first learn a temporary model for the current task, and then pseudo labels for the unlabeled data are computed by fusing information from the old model of the previous task and the current temporary model. Additionally, conflict reduction is proposed to resolve the conflicts of pseudo labels generated from both the old and temporary models. We show that maximizing self-entropy can further improve results by smoothing the overconfident predictions. Interestingly, in the experiments we show that the auxiliary data can be different from the training data and that even general-purpose but diverse auxiliary data can lead to large performance gains. The experiments demonstrate state-of-the-art results: obtaining a relative gain of up to 114\%  on Pascal-VOC 2012 and 8.5\% on the more challenging ADE20K compared to previous state-of-the-art methods. 
\end{abstract}


\IEEEpeerreviewmaketitle

\section{Introduction}
\IEEEPARstart{S}{emantic} segmentation is a fundamental research field in computer vision. It aims to predict the class of each pixel in a given image. The availability of large labeled datasets~\cite{everingham2011pascal,zhou2017scene} and the development of deep neural networks~\cite{minaee2020image} have resulted in significant advancements. The vast majority of semantic segmentation research focuses on the scenario where all data is jointly available for training.
However, for many real-world applications, this might not be the case, and it would be necessary to incrementally learn the semantic segmentation model. Examples include scenarios where the learner has only limited memory and cannot store all data (common in robotics), or where privacy policies might prevent the sharing of data (common in health care)~\cite{de2019continual,parisi2019continual}.

Incremental learning aims to mitigate catastrophic forgetting~\cite{mccloskey1989catastrophic} which occurs when neural networks update their parameters for new tasks. Most work has focused on image classification~\cite{de2019continual,parisi2019continual,kemker2018measuring}, while less attention has been dedicated to other applications, such as object detection~\cite{shmelkov2017incremental} and semantic segmentation~\cite{tasar2019incremental,michieli2019incremental,cermelli2020modeling}. Recently MiB~\cite{cermelli2020modeling} achieved state-of-the-art results on incremental semantic segmentation. Its main novelty is to model the background distribution shift during each incremental training session by reformulating the conventional distillation loss. However, the forgetting of learned knowledge is still severe due to the lack of previous labeled data.

\begin{figure}[tb]
\begin{center}
\includegraphics[width=0.5\textwidth]{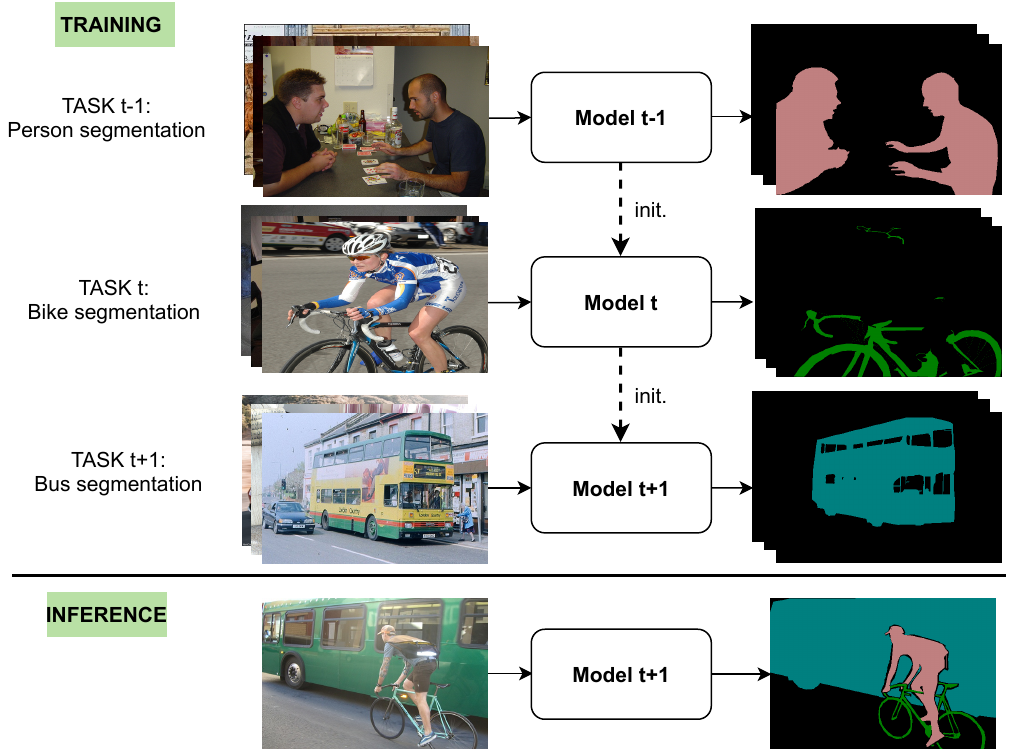}
\caption{Illustration of class-incremental semantic segmentation. 
During the training phase, at each task we only get the ground truth labels of one class (or a few classes). The background may contain objects from previous tasks, e.g. person from the previous task t-1 is annotated as background at task t. During the inference phase, pixels are required to be segmented into all classes including person, bike and bus.
}\label{fig:ss}
\end{center}
\end{figure}

\figintro

\figconflict  

An illustration of class-incremental semantic segmentation is provided in Fig.~\ref{fig:ss}. In this example, at task t we have ground truth annotation (bike and background), where the background contains person (from the previous task t-1). The model is required to learn continually and segment all seen objects during inference time, including person, bike and bus.  The main challenge of incremental semantic segmentation  is the inaccessibility of previous data. We here consider the scenario where no data of previous tasks can be stored, as is common in incremental semantic segmentation~\cite{michieli2019incremental,cermelli2020modeling}. Storing of data could be prohibited due to privacy concerns or government regulations, and is one of the settings studied in class-incremental learning~\cite{parisi2019continual}. To address this problem, we propose to use self-training to exploit unlabeled auxiliary data. Self-training~\cite{nigam2000analyzing,zou2018unsupervised,zou2019confidence} has been successfully applied to semi-supervised learning and domain adaptation. It aims to first predict pseudo labels of the unlabeled data and then learn from them iteratively. To the best of our knowledge, self-training has not yet been explored for incremental learning.

In this work, we propose a self-training framework for incremental semantic segmentation, as shown in Fig.~\ref{fig:intro}. The idea is to introduce self-training of unlabeled data to aid the incremental semantic segmentation task. Specifically, we first train a temporary model with current labeled data, then pseudo labels are predicted and fused for auxiliary unlabeled data by both the old and temporary models. We retrain a new model on auxiliary data to mitigate catastrophic forgetting. Simply fusing the pseudo labels from two models causes problems due to conflicting predictions.
We show some challenges in Fig.~\ref{fig:v-conflict} for generating pseudo labels for unlabeled data. 
Fusing the label information from both models is not straightforward. It is clear that neither the prediction from the old model (second column) or the temporary model (third column) is ideal. Therefore, we further propose a conflict reduction mechanism in Section~\ref{sec:method} to fuse pseudo labels (last column) to learn a new model. Additionally, predicted pseudo labels from neural networks are often over-confident~\cite{zou2019confidence,pereyra2017regularizing}, which might mislead the training.  We therefore propose to maximize self-entropy to smooth the predicted distribution and reduce the confidence of predictions.   

Our main contributions are:

\begin{itemize}
    \item We are the first to apply self-training for class-incremental semantic segmentation to mitigate forgetting by rehearsal of previous knowledge using auxiliary unlabeled data.  
    \item We propose a conflict reduction mechanism to tackle the conflict problem when fusing the pseudo labels from the old and temporary models for auxiliary data. 
    \item We show that maximizing the self-entropy loss can smooth the overconfident predictions and further improve performance. 
    \item We demonstrate state-of-the-art results, obtaining up to 114\%  relative gain on Pascal-VOC 2012 dataset and 8.5\% on the challenging ADE20K dataset compared to the MiB method.
\end{itemize}

\section{Related Work}

\subsubsection{Semantic Segmentation}
Image segmentation has achieved significant improvements with the advance of deep neural networks~\cite{minaee2020image}. 
Fully Convolutional Networks (FCNs)~\cite{long2015fully} is one of the first works for semantic segmentation to use only convolutional layers, and can take any arbitrary size of input images and output segmentation maps. Encoder-Decoder based architectures are popular for semantic segmentation. The deconvolutional (transposed convolutional) layer~\cite{noh2015learning} is proposed to generate accurate segmentation maps. SegNet~\cite{badrinarayanan2017segnet} proposes to use indices of encoder max-pooling to upsample the corresponding decoder. Additionally, multi-resolution information~\cite{zhao2017pyramid,he2019adaptive}, attention
mechanism~\cite{chen2016attention,fu2019dual} and dilated convolution (atrous convolution)~\cite{chen2017deeplab,chen2017rethinking,chen2018encoder}  are further developed to improve performance.
Apart from single-modal data based semantic segmentation methods, multi-modal data fusion-based methods were also proposed by incorporating other input modalities leading to further improvements. RTFNet~\cite{sun2019rtfnet} fuses both RGB and thermal information to perform semantic segmentation for autonomous vehicles. DFM~\cite{wang2021dynamic} developed a benchmark of existing data-fusion networks evaluating the fusion of different types of visual features. \cite{hazirbas2016fusenet, wang2018depth, chen2020bi}, incorporate depth into semantic segmentation via a fusion-based architecture. RoadSeg~\cite{fan2020sne} fuses features from both RGB images and the inferred surface normal information for freespace detection.

However, these works assume a static world to learn semantic segmentation with all data available. While our method
considers a more realistic setting, where a model has to adapt continually to new tasks.

\subsubsection{Incremental Learning}
The problem of catastrophic forgetting~\cite{mccloskey1989catastrophic} has been studied extensively in recent years when neural networks are required to adapt to new tasks. Most work has been focused on image classification. It can be roughly divided into three categories according to~\cite{de2019continual,parisi2019continual,kemker2018measuring}: Regularization-based~\cite{li2017learning,kirkpatrick2017overcoming,zenke2017continual,jung2018less, aljundi2018memory}, rehearsal-based~\cite{rebuffi2017icarl,chaudhry2018riemannian} and architecture-based methods~\cite{rusu2016progressive,mallya2018piggyback,mallya2018packnet}. Regularization-based methods alleviate  forgetting of previously learned knowledge by introducing additional regularization term to constraint the output embeddings or parameters while training the current task. Knowledge distillation has been very popular for several methods~\cite{li2017learning,castro2018end, hou2019learning}.
Rehearsal-based methods usually need to store exemplars (small amounts of data) from the previous tasks which are used to be replayed. Some approaches propose alternative ways of replay to avoid storing exemplars, including using a generative model to do rehearsal. Architecture-based methods dynamically grow the network to increase capacity to learn new tasks, while the old part of the network can be protected from forgetting.

Due to privacy issues or memory limits, it is not always possible to access data from previous tasks, which causes catastrophic forgetting of previous knowledge. In this paper, we consider this more difficult setting of \emph{exemplar-free class-IL} in which the storing of previous task data is prohibited. Unlabeled data is seen as an alternative to secure privacy and mitigate forgetting. There are some works that employ unlabeled data in the context of the continual learning of an image classification system. Zhang et al. ~\cite{zhang2020class} propose to train a separate model with only new data and then use auxiliary data to train a student model using a distillation loss with both new and old models. Lee et al. ~\cite{lee2019overcoming}  propose confidence-based sampling to build an external dataset. While there is no existing work on pixel-wise incremental task leveraging unlabeled data.

Recently, the attention of continual learning has also moved to other applications, such as object detection~\cite{shmelkov2017incremental}, and semantic segmentation~\cite{michieli2019incremental,cermelli2020modeling}. Shmelkov et al.~\cite{shmelkov2017incremental} propose to use a distillation loss on both bounding box regression and classification outputs for object detection. A incremental few-shot detection (iFSD) setting is proposed in~\cite{perez2020incremental}, where new classes must be
learned incrementally (without revisiting base classes)
and with few examples.  Michieli et al.~\cite{michieli2019incremental} propose to use distillation both on the output logits and intermediate features for incremental semantic segmentation. Recently,  MiB (Cermelli et al.~\cite{cermelli2020modeling}) achieves state-of-the-art performance by considering previous classes  as background for the current task and current classes as background for distillation. It is also investigated in remote sensing~\cite{tasar2019incremental} and medical data~\cite{ozdemir2018learn} for incremental semantic segmentation. 
In knowledge distillation one trains the new task while distilling the knowledge of the previous model; as a consequence the model might have suboptimal performance on the current task. In our approach,we allow the new model to adapt totally to the new task, and then in the second phase we aim to join the knowledge of both the old model and the current temporary model in a new model.

\begin{figure*}[tb]
\begin{center}
\includegraphics[width=1\textwidth]{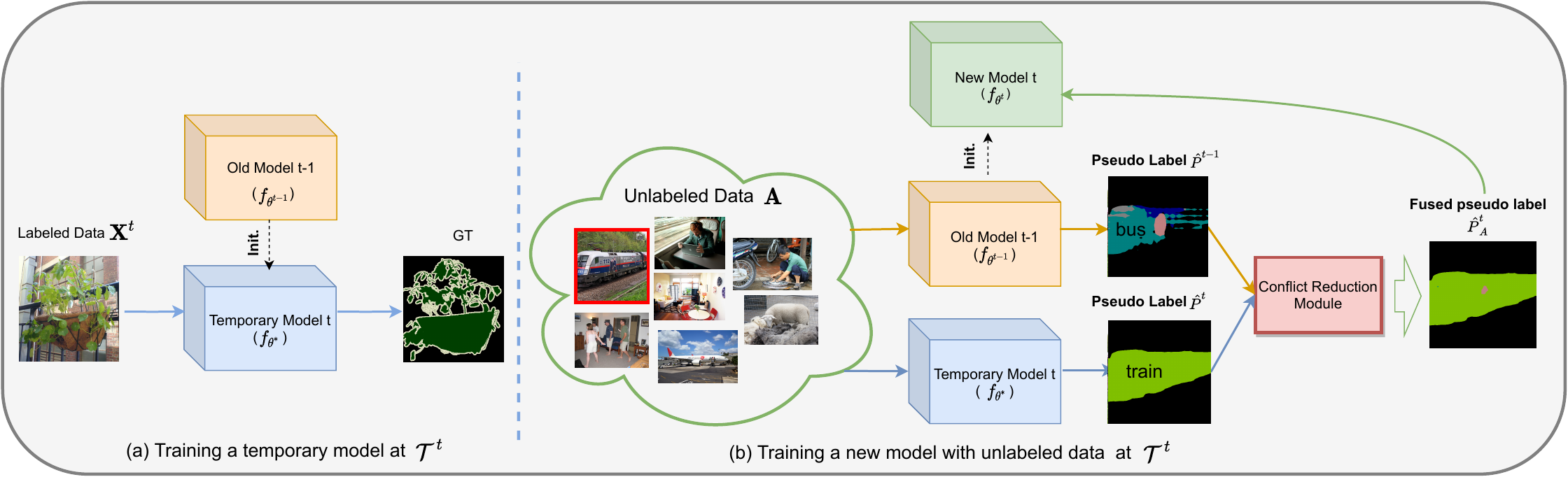}
\caption{Overview of our method. (a) A temporary model is initialized with the old model from the previous session and learned at the current session $\mathcal{T}^t$ with labeled data. (b) Pseudo labels are generated and fused by leveraging unlabeled data, where a conflict reduction module is proposed to generate more accurate pseudo labels to learn a new model. As an example in (b), the category ‘train’ is not learned in previous training sessions, therefore it is most likely to be predicted as a similar category ‘bus’ (top segmentation map). After learning ‘train’ on the current task, it is predicted as ‘train’ correctly (bottom segmentation map). To generate more accurate pseudo labels, conflict reduction is proposed to fuse the two predictions (right segmentation map).}
\label{fig:framework}
\end{center}
\end{figure*}

\subsubsection{Self-Training}
Self-training~\cite{nigam2000analyzing,grandvalet2005semi,lee2013pseudo} aims to leverage unlabeled data by computing pseudo labels for it with a teacher model trained on existing labeled data. Self-training iteratively generates one-hot pseudo-labels corresponding to the prediction confidence of a teacher model, and then retrains a network based on these pseudo-labels. Wang et al.~\cite{wang2019self} designed a traditional method to generate coarse labels (pseudo labels), and then used the coarse labels to train existing semantic segmentation networks to achieve results better than traditional methods. Recently, it has achieved significant success on semi-supervised learning~\cite{grandvalet2005semi,triguero2015self} and  domain adaptation~\cite{zou2018unsupervised, chen2011co}. However, the predicted pseudo labels tend to be over-confident, which might mislead the training process and hurt the learning behaviour~\cite{bagherinezhad2018label}. Different methods to solve noisy label learning such as label smoothing~\cite{pereyra2017regularizing} and confidence regularization~\cite{zou2018unsupervised,zou2019confidence} are proposed to mitigate this phenomena. In this work, we explore self-training for learning semantic segmentation sequentially with confidence regularization. Moreover, a conflict reduction mechanism is proposed to fuse the pseudo labels specifically for incremental learning, which is different from ensemble networks~\cite{kumar2016ensemble,fan2020ensemble}, where different models are complementary to each other and there are no conflicts between them.

\section{Proposed Method}
\label{sec:method}

\subsection{Class-Incremental Semantic Segmentation} 

Semantic segmentation aims to assign each pixel $(x_i,x_j)(1\leq i\leq  h, 1 \leq j  \leq  w)$ of image $\mathbf{x}$ a label 
$y_{i,j} \in \mathbfcal{Y}, \mathbfcal{Y}\in \{0, 1, ..., N-1\}$, representing the semantic class. Here,
$h$ and $w$ are the height and width of the input image $\mathbf{x}$, 
$N$ is the number of classes, and we define class $0$ to be the background. 
The setting for class incremental learning (CIL) for semantic segmentation was first defined by 
\cite{michieli2019incremental,cermelli2020modeling}.  Training is conducted for CIL along $T$ different training sessions. During each training session, we only have training data of newly available classes, while the training data of the previously learned classes are no longer accessible. Each session introduces \emph{novel categories} to be learned. 

Specifically, the $t_{th}$ training session contains data $\mathbfcal{T}^t= (\mathbf{X}^t, \mathbf{Y}^t)$, where $\mathbf{X}^t$ contains the input images for the current task and $\mathbf{Y}^t$ is the corresponding ground truth. The current label set $\mathbfcal{Y}^{t}$ is the combination of  the previous label set $\mathbfcal{Y}^{t-1}$ and a set of new classes $\mathbf{C}^{t}$, such that $\mathbfcal{Y}^{t}=\mathbfcal{Y}^{t-1} \cup \mathbf{C}^t$. Only pixels of new classes are annotated in $\mathbf{X}^t$ and the remaining pixels are assigned as background. The loss function for the current training session is defined as follows: 
\begin{equation}
   \label{eq:obj-general}
    \mathcal{L}_{ce}(\theta^t)= \frac{1}{|\mathbfcal{T}^t| 
    }\sum_{(\mathbf{x},y)\in\mathbfcal{T}^t}
    \ell_{ce}({\theta^t};\mathbf{x},y) 
\end{equation}
where $\ell_{ce}$ is the standard cross-entropy loss used for supervised semantic segmentation and $|\mathbfcal{T}^t|$ is total number of samples in current training session $\mathbfcal{T}^t$.

\subsection{Self-Training for Incremental Learning}

A naive approach to address the class-incremental learning (CIL) problem is to train a model $f_{\theta^t}$ on each set $\mathbf{X}^{t}$ sequentially by simply fine-tuning $f_{\theta^t}$ from the previous model $f_{\theta^{t-1}}$. This approach would suffer from catastrophic forgetting, as the parameters of the model are biased to the current categories because no samples from previous data $\mathbf{X}^{t-1}$ are replayed. As discussed in the related work, various approaches to prevent forgetting could be considered, like regularization~\cite{li2017learning,kirkpatrick2017overcoming,zenke2017continual,jung2018less, aljundi2018memory}, or rehearsal methods~\cite{rebuffi2017icarl,chaudhry2018riemannian}. Instead, we propose to use self-training. To the best of our knowledge, we are the first to investigate self-training for incremental learning. Our goal is to use unlabeled data to generate \emph{pseudo labels} for replay. In this way, the models are able to ‘revisit’ the previous knowledge and avoid catastrophic forgetting of previously learned categories. 

We present an overview of our framework on incremental semantic segmentation with a self-training mechanism shown in Fig.~\ref{fig:framework}. There are two training steps for each task. We first initialize a temporary model $f_{\theta^*}$ from the old model $f_{\theta^{t-1}}$ (trained in the training session $\mathbfcal{T}^{t-1}$) and update parameters using $\mathbf{X}^{t}$ in training session $\mathbfcal{T}^t$. This temporary model is trained to be optimal for the new classes $\mathbf{C}^t$, however it forgets the previous classes $\mathbf{Y}^{t-1}$. To overcome this problem, during the second step, we combine the knowledge from both the old and temporary models in order to predict all categories we have seen.
We generate pseudo label $\hat{P}^{t-1}$ and $\hat{P}^{t}$ by feeding the unlabeled data $\mathbf{A}$ to the previous model $f_{\theta^{t-1}}$ and the current temporary model $f_{\theta^*}$, respectively. The generated pseudo labels of the auxiliary data from both models have the potential to represent all categories the models have encountered. 

 We require a strategy to fuse the pseudo-labels of both models for each image in the auxiliary dataset into a single pseudo-labeled image. We first propose a fusion based on the idea that the predictions from the previous model $\hat{P}^{t-1}$ should be trusted and we only change those background pixels that are considered foreground in the current temporary model $\hat{P}^t$. In Section~\ref{sec:conflict}, we improve this fusion of pseudo labels by considering a conflict reduction mechanism. 

When $\hat{P}^{t-1}$ and $\hat{P}^{t}$ both consider the pixel as background, we assign background label 0 to the corresponding pixel in the fused pseudo label $\hat{P}^{t}_A$ directly; when $\hat{P}^{t-1}$ consider the pixel as background (label is 0), while $\hat{P}^{t}$ classify it as foreground (label is larger than 0), $\hat{P}^{t}_{\mathbf{A}}$ is equal to $\hat{P}^t$ since the pixel is likely  to be the category learned in the current temporary model $f_{\theta^t}$; If $f_{\theta^{t-1}}$ considers the pixel as foreground, $\hat{P}^{t}_{\mathbf{A}}$ is equal to $\hat{P}^{t-1}$, we assume the old model has higher priority than the current temporary model since the previous model is accumulating more and more knowledge. The final fusion of pseudo labels for these auxiliary data $\hat{P}^{t}_{\mathbf{A}}$ can be presented as follows:

\begin{equation} 
\hat{P}^{t}_{\mathbf{A}}=
\begin{cases}
0 & \text{ if }  \hat{P}^t =0  \text{ and }  \hat{P}^{t-1}=0 \\
\hat{P}^t & \text{ if } \hat{P}^t>0  \text{ and }  \hat{P}^{t-1}=0 \\
\hat{P}^{t-1}& \text{ if } \hat{P}^{t-1} >0
\end{cases}
\label{eq:pseudo1}
\end{equation}

Finally, we update the new model (initialized from the old model $f_{\theta^{t-1}}$) using the auxiliary data $\mathbf{A}$ and pseudo label $\hat{P}^{t}_{\mathbf{A}}$. We repeat the above procedure until all tasks are learned. 
For many practical applications, the assumption of an available auxiliary dataset is realistic. For instance, in the application of autonomous driving, there is abundant amount of unlabeled data available for training. In our experiments, we will show results where we use the COCO dataset as the auxiliary dataset for Pascal-VOC 2012, and the Places365 for ADE20K in Section~\ref{sec:exp} . We also investigated how unrelated data affects the self-training process.

\subsection{Conflict Reduction}\label{sec:conflict}

In the above section, we introduce how self-training is adapted in the framework for incremental semantic segmentation to help mitigate catastrophic forgetting of old categories. The pseudo labels of the auxiliary data are generated from the old and temporary model, respectively. Then the two pseudo labels are fused into the final pseudo label directly, however, there might be wrong fusions due to similar categories that are often mis-classified (see Fig. 2). Therefore, we propose conflict reduction to further improve the accuracy of pseudo label fusion. 

Assume ‘bus’ is the category added in the training session $\mathbfcal{T}^{t-s}$ $(s\geq 1)$ , ‘train’ is learned in the current training session $\mathbfcal{T}^t$ (as seen in Fig.~\ref{fig:framework}). When an image from the auxiliary data containing ‘train’  is fed into the model $f_{\theta^{t-1}}$, as ‘train’ has never been learned in the previous training sessions, the model assigns the maximum probability to the most similar category label ‘bus’ due to the usage of cross entropy loss. Following Eq.\ref{eq:pseudo1}, the fused pseudo labels are automatically assigned from $\hat{P}^{t-1}$ if the previous model regards the pixels as foreground with no need to check the pseudo label $\hat{P}^t$ obtained from the current temporary model. This results in the mis-classification and a drop in performance of the overall semantic segmentation system. In this case the current temporary model $f_{\theta^*}$ labels the train as ‘train’ very confidently, since it just learned to recognize trains from data $\mathbf{X}^t$. Conflict frequently occurs when similar categories appear such as ‘sheep’ and ‘cow’, ‘sofa’ and ‘chair’, ‘bus’ and ‘train’ et al.. Therefore, a Conflict Reduction module is proposed to obtain better fusion when both the old and temporary models consider the pixel as foreground ( $\hat{P}^t>0$ and $\hat{P}^{t-1}>0$). Therefore, we update the Eq.\ref{eq:pseudo1} as follows:

\begin{equation} 
\label{eq:conflict}
\hat{P}^{t}_{\mathbf{A}}=
\begin{cases}
0 & \text{ if }  \hat{P}^t =0  \text{ and }  \hat{P}^{t-1}=0 \\
\hat{P}^t & \text{ if } \hat{P}^t >0  \text{ and }  \hat{P}^{t-1}=0 \\
\hat{P}^{t-1} & \text{ if } \hat{P}^t=0  \text{ and }  \hat{P}^{t-1}>0 \\

\hat{P}^t & \text{ if } \hat{P}^t, \hat{P}^{t-1} >0  \text{ and }  \text{max} (\hat{\mathbf{q}}^{t}) > \text{max} (\hat{\mathbf{q}}^{t-1} ) \\

\hat{P}^{t-1} & \text{ if } \hat{P}^t, \hat{P}^{t-1} >0  \text{ and }  \text{max} (\hat{\mathbf{q}}^{t}) <\text{max} (\hat{\mathbf{q}}^{t-1}) \\

\end{cases}
\end{equation}
where $\hat{\mathbf{q}}^{t-1}$ and $\hat{\mathbf{q}}^{t}$ are output probabilities from the old and temporary models, respectively.

\subsection{Maximizing Self-Entropy}
Pseudo-labeling is a simple yet effective technique commonly used in self-training learning. One major concern of training with pseudo label is the lack of guarantee for label correctness~\cite{zou2019confidence}. To address this problem, maximizing the self-entropy loss is explored in this work to relax the pseudo labels and redistribute a certain amount of confidence to other classes. We soften the pseudo label by maximizing the self-entropy loss $\mathcal{L}_{se}(\theta^t)$ according to: 

\begin{equation}
\label{eq:total}
    \mathcal{L}(\theta^t) = \mathcal{L}_{ce}(\theta^t) - \lambda * \mathcal{L}_{se}(\theta^t)
\end{equation}
where,
\begin{equation}
    \mathcal{L}_{se}(\theta^t)= - \frac{1}{|\mathbfcal{T}^t| 
    }\sum_{}
    \hat{\mathbf{q}}\log \hat{\mathbf{q}}
\end{equation}
Note that self-entropy loss is applied in different stages of training besides when learning the new model from pseudo labels. For incremental learning, the models are updated for different tasks, the current new model will become the old model for the next task in the future. Therefore reducing the overconfident predictions at all stages is crucial for generating more correct  pseudo labels for incremental learning.

\subsection{Algorithm}
We provide a detailed algorithm of our incremental semantic segmentation procedure in Algorithm~\ref{table:algorithm}. For the first task (line 5), it is similar to standard incremental learning methods. For the remaining tasks, our method can be divided into two main parts: pseudo label fusion (lines 7-9) and model retraining (line 10).

\tablealgorithm  

\section{Experiments}
\label{sec:exp}

\tablepascal  

\subsection{Experimental setups}

In this section, we provide details for the datasets, evaluation metrics, implementations and compared methods. Code will be made available upon acceptance of this manuscript.

\subsubsection{Datasets} We evaluate all methods using Pascal-VOC 2012 and ADE20K. Pascal-VOC 2012~\cite{everingham2011pascal} has 10,582 images for training, 1,449 images for validation and 1,456 images for testing. Images contain 20 foreground object classes and one background class. ADE20K~\cite{zhou2017scene} is a large scale dataset containing 20,210 images in the training set, 2,000 images in the validation set, and 3,000 images in the testing set. It contains 150 classes of both stuff and objects. For auxiliary datasets, we choose COCO 2017 train set~\cite{lin2014microsoft} with 80 classes and 118K images for Pascal-VOC 2012 and Places365~\cite{zhou2017places} with 365 classes and 1.8M images for ADE20K. 

\subsubsection{Implementation Details} We follow the same implementation as proposed in~\cite{cermelli2020modeling}. For all methods, we use the same framework in Pytorch. The Deeplab-v3 architecture~\cite{chen2017rethinking} with ResNet-101~\cite{he2016deep} backbone is used for all methods. In-place activated batch normalization is used and the backbone network is initilized with an ImageNet pretrained model~\cite{rota2018place}. We train the network with SGD and an initial learning rate of $10^{-2}$ for the first task and $10^{-3}$ for the rest of the tasks as in~\cite{cermelli2020modeling}. We train the current model on Pascal-VOC 2012 for 30 epochs and on ADE20K for 60 epochs.We train the model with a batch size of 24. And we crop the images to 512 × 512 during both training and test phases. The self-training procedure is trained on unlabeled data for one pass, the trade-off $\lambda$ between cross entropy and self-entropy is set to 1.  20\% of training data are used as validation set and the final results are on the standard test set. 

\subsubsection{Compared Methods} We consider the Fine-tuning (FT) baseline and the Joint Training (Joint) upper bound. Additionally, we compare with several regularization-based methods adapted to incremental semantic segmentation including ILT~\cite{michieli2019incremental}, LwF~\cite{li2017learning},  LwF-MC~\cite{rebuffi2017icarl},
RW~\cite{chaudhry2018riemannian},
EWC~\cite{kirkpatrick2017overcoming} and
PI~\cite{zenke2017continual}. We also compare with state-of-the-art method MiB~\cite{cermelli2020modeling}. Additionally, MiB can be further trained with unlabeled data by generating pseudo labels with learned current model, we denote it as MiB + Aux by leveraging unlabeled data using pseudo labels.  All results are reported in mean Intersection-over-Union (mIoU) in percentage. It is averaged for all classes of each task after learning all tasks. Note that none of the methods has access to exemplars from previous tasks.

\subsection{On Pascal-VOC 2012}
We compare different methods in three different scenarios as in~\cite{cermelli2020modeling}. 19-1 means we first learn a model with first 19 classes and then learn the remaining class as the second task. For 15-5 scenario, there are 15 classes for the first task and the remaining 5 classes as the second task. For 15-1, the first task is the same as in 15-5 scenario, but the remaining 5 classes are learned one-by-one resulting in a total of six tasks. For all scenarios, we consider two different settings. \textit{Disjoint} setting assumes images of current task only contain the current or previous classes. While \textit{Overlapped} setting assumes that at each training session, it contains all images with at least one pixel of novel classes. Previous classes are labeled as background for both settings.

\figvoc 

\begin{figure*}[tb]
\begin{center}
\includegraphics[width=0.32\textwidth]{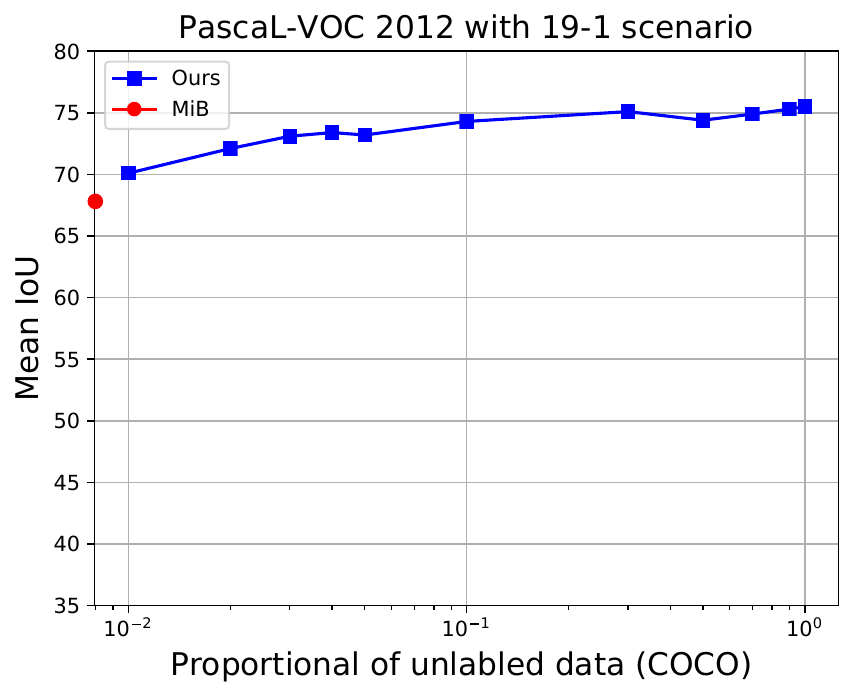}
\includegraphics[width=0.32\textwidth]{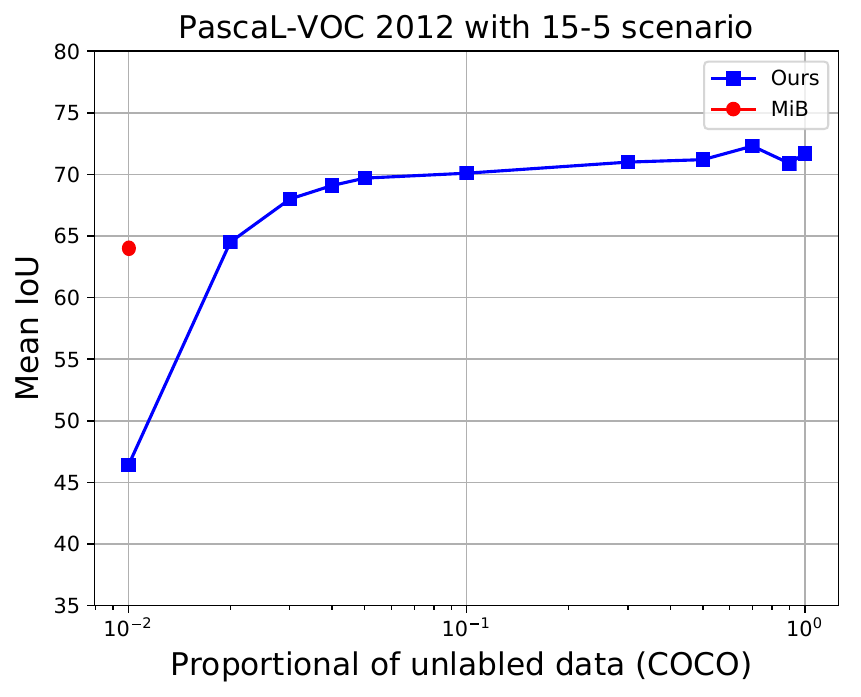}
\includegraphics[width=0.32\textwidth]{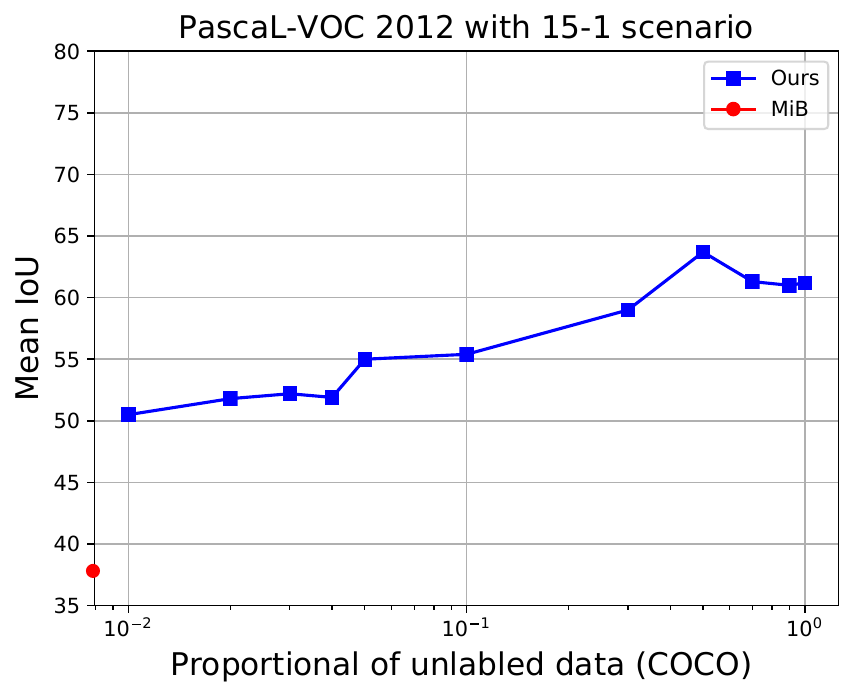}
\caption{Mean IoU on Pascal VOC 2012 for 19-1, 15-5 and 15-1  scenarios as a function of proportional of unlabeled data. The curve starts from 1\% (about 1K images) and ends at 100\%. The horizontal axis is in logarithmic scale. }
\label{fig:datasize}
\end{center}
\end{figure*}

\subsubsection{Addition of one class (19-1)} As shown in Table ~\ref{tab:pascal}, in this scenario, FT and PI obtain the worst results, where they forget almost all of the first task, and perform poorly on the new task with 6.2\% and 5.9\% of mIoU in the \textit{Disjoint} setting, respectively. EWC and RW, both weights-based regularization methods, improve quite a lot compared to PI. Interestingly, activation-based regularization methods LwF, LwF-MC and ILT perform significant better in both \textit{Disjoint} and \textit{Overlapped} settings. MiB is superior to all previous methods but inferior to our method by a large margin. Our method surpasses MiB on overall mIoU by 8.0\% for the \textit{Disjoint} setting and 6.7\% for the \textit{Overlapped} setting. When MiB is further trained with unlabeled data (MiB + Aux) using generated pseudo labels, it improves by 2.2\% for the \textit{Disjoint} setting and 11.3\% for the \textit{Overlapped} setting on class 20, while keeping similar performance for the first 19 classes. Note that our method is very close to Joint training performance on both settings (75.4\% to 77.4\% and 74.5\% to 77.4\%).

\subsubsection{Single-step addition of five classes (15-5)} Similar conclusions can be drawn in this scenario. Our method outperforms MiB by 6.6\% in the \textit{Disjoint} setting and 2.1\% in the \textit{Overlapped} setting.
The performance gain due to additional pseudo labels for MiB+Aux is more obvious in this scenario (when compared to the 19-1 setting).
However, there is  still a big gap compared to our method. This shows that our proposed techniques are more efficient to leverage knowledge from unlabeled data.  Our method achieves similar overall results (71.3\% and 71.1\%) in both settings, showing the robustness of our proposed method. 

We report some qualitative results for different incremental methods (Ours, MiB and Fine-tuning) on Pascal-VOC 2012 with 15-5 scenario in Fig.~\ref{fig:voc}. The results demonstrate the superiority of our approach. FT totally forgets previously learned classes (first row and third row) but correctly predicts new classes (second row),  while our approach obtains sharper (e.g. person, bike), more coherent (e.g. potted plant) and finer-border (e.g. cow) predictions compared to the state-of-the-art method MiB. 

\subsubsection{Multi-step addition of five classes (15-1)} This is the most challenging scenario of the three. There are five more tasks after learning the first task, therefore, it is more difficult to prevent forgetting. From Table~\ref{tab:pascal}, we can observe that in general all methods forget more in this scenario. MiB only achieves 46.2\% and 35.1\% for the first task in two settings after learning all tasks. While our method achieves 70.1\% and 71.4\% of mIoU. Meanwhile our method outperforms MiB on new tasks by a large margin (21.4\% and 26.5\% respectively) as well. Overall, we gain 23.3\% for the \textit{Disjoint} setting and 33.9\% for the \textit{Overlapped} setting. We also see a significant improvement for MiB + Aux compared to the original MiB, from 37.9\% to 39.9\% for \textit{Disjoint} and from 29.7\% to 37.3\% for the \textit{Overlapped} setting. Again, without any more expensive annotation processes, unlabeled data is beneficial not only for our proposed method, but also very useful for the state-of-the-art method MiB to further boost performance.

\subsection{Ablation study}\label{sec:ablation}

In this section, we perform an ablation study on several different aspects including the proposed techniques, the relationship between performance and the amount of unlabeled data used, epochs for self-training and hyper-parameter $\lambda$.

\subsubsection{Impact of different proposed components} 
In Table~\ref{tab:ablation} we conduct an ablation study on the Pascal-VOC 2012 \textit{Disjoint} setting. In the first row, we show the performance of FT on three scenarios. And in the second row, MiB is the current state-of-the-art method. When we use self-training ($ST$) with unlabeled data (Eq.2), the performance is largely improved in all three scenarios, which shows the effectiveness of self-training on unlabeled data in incremental semantic segmentation.  We also conducted experiments for self training, assuming the temporary model has higher priority than the old model for unlabeled data (denoted as $ST^{*}$) as an ablation study. The results are much worse than $ST$, which shows our original assumption results in much better performance. 
Conflict reduction ($ST+CR$) further improves the results on all of these three scenarios by 0.6\%, 4.2\% and 3.2\%, respectively. Maximizing self-entropy ($ST+CR,MS$) strategy further obtains 0.4\% better on the 19-1 scenario and  0.8\% better results on the 15-5 scenario. On more challenging 15-1 scenario, MS boosts performance significantly by 2.4\%.
Note that the gain for the 19-1 scenario is small compared to the other two scenarios because there is only one category (‘TV’) for the second task, it does not contribute much to the overall performance even without learning it. Therefore, for incremental learning, the other two scenarios are more relevant.

\tableablation 

\begin{table}[tb]
\begin{center}
\caption{Mean IoU on the Pascal VOC 2012 by using 1\% of unlabeled data (COCO) with different number of epochs. \textit{Disjoint} setting is used in this experiment.}
\resizebox{0.37\textwidth}{!}{%
\begin{tabular}{c||cccc}
Epochs & 1 & 5 & 10 & 20 \\ \hline
19-1 & 70.1 & 72.8 & 73.2 & 73.1 \\
15-5 & 46.4 & 66.9 & 68.3 & 68.8 \\ 
15-1 & 50.5 & 54.1 & 56.9 & 58.1 
\label{table:epoch}
\end{tabular}}
\end{center}
\end{table}

\subsubsection{The amount of unlabeled data} We also evaluate the relationship between mean IoU and the size of unlabeled data by randomly selecting a portion of unlabeled data (see Fig.~\ref{fig:datasize}). We experiment on \textit{Disjoint} setup for 19-1, 15-5 and 15-1 scenarios. Notably, for 19-1 scenario, our method beats MiB by using only 1\% unlabeled data, and the mIoU continually increases when more unlabeled data is used. For 15-5 scenario, our method achieves similar results as MiB by only using 2\% of unlabeled data. It keeps improving by increasing unlabeled data and peaks when 70\% of unlabeled data are used. Similar conclusion can be observed for the 15-1 scenario, our method outperforms MiB by a large margin with only 1\% unlabeled data. When adding more unlabeled data, the curve goes up consistently until it reaches the best performance and then it drops a bit in the end. 

\subsubsection{Number of self-training epochs} Without any specific mention, we only pass all unlabeled data once into the network for  efficiency consideration throughout the paper in our experiments. In this section, we consider different multiple passes when only 1\% unlabeled data is available from COCO dataset as shown in Table ~\ref{table:epoch}. As expected, compared to only passing it once, training for more epochs achieves significant gain for all three scenarios. Specifically, when we increase the self-training epochs from 1 to 20, it improves from 70.1\% to 73.1 for 19-1 scenario, from 46.4 to 68.8 for 15-5 scenario and from 50.5 to 58.1 for 15-1 scenario, respectively. It is the most effective to 15-5 scenario because the difficulty of this scenario is between the other two scenarios. It also shows that by training multiple runs using pseudo labels the performance can be further boosted even in the low-data regime, when we have little unlabeled data.

\subsubsection{Impact of trade-off $\lambda$} This parameter controls the strength between the cross-entropy and self-entropy loss.  
 In this section, we report results using various $\lambda$'s on the 15-5 scenario of the \textit{Disjoint} setup. As shown in Table~\ref{tab:lambda}, by changing $\lambda$ from 0.1 to 5, the overall performance goes up first and then goes down. When 0.5 and 1 are used, it obtains the best performance. Using $\lambda=1$ has similar performance as 0.5 and slightly better on the new task, without any specific mention,  $\lambda=1$ is used throughout the paper.

\subsubsection{Comparability of two output probabilities} To generate pseudo labels for self-training, we fuse labels from the old model and the temporary model by comparing their output probabilities. In order to show whether the output probabilities from two different models are comparable, we conduct an ablation study by adding a bias on the probability of the old model before comparing these two. As shown in Table~\ref{tab:bias}, our method performs the best with $bias=0$, which is adopted throughout the paper by default. When we increase or decrease the bias, the performance drops significantly. It shows that it is reasonable to compare the probabilities of different models.

\begin{table}[tb]
\begin{center}
\caption{The trade-off $\lambda$ between cross entropy and self-entropy on 15-5 scenario. Mean IoU is reported for both tasks and overall performance is in the last row.}
\resizebox{0.4\textwidth}{!}{%
\begin{tabular}{c||cccc}
Trade-off $\lambda$ & 0.1  & 0.5   & 1   & 5            \\ \hline
1-15            & 76.8 & 77.4   &  77.1   & 72.4    \\
16-20              &  53.1 &  55.4  & 55.6    &   49.7         \\
all                  & 
70.9
    &  71.9 & 71.7   &       66.7     \\           
\end{tabular}}
\label{tab:lambda}
\end{center}
\end{table}

\begin{table}[tb]
\begin{center}
\caption{Comparability of two output probabilities on 15-5 scenario with different bias. Mean IoU is reported for both tasks and overall performance is in the last row.}
\resizebox{0.4\textwidth}{!}{%
\begin{tabular}{c||ccccc}
Bias & 0.1  & 0.05   & 0   & -0.05 & -0.1            \\ \hline
1-15            & 75.5 & 76.0  & \textbf{76.9} & 73.4 & 70.3 \\
16-20              & 50.0  & 53.4  & \textbf{54.3}   & 50.7  & 47.3        \\
all                 &  69.1 & 70.4 & \textbf{71.3}  & 67.7  & 64.5 \\           
\end{tabular}}
\label{tab:bias}
\end{center}
\end{table}

\subsection{On different self-training datasets}

In this section, we compare several datasets used as self-training datasets for Pascal-VOC to show how sensitive our method is with respect to the unlabeled data.
We choose a general-purpose ImageNet validation set, two fine-grained CUB\_200\_2011 (Birds) and Flowers datasets together with the COCO dataset as different unlabeled data to show how our method performs on Pascal-VOC 15-5 scenario (see Table~\ref{tab:fid}). It is surprising that using ImageNet (validation set) provides similar performance as using COCO dataset (mIoU: 70.0 vs 71.3) for all classes. It suggests that datasets with diverse categories can be a good option. As expected, it fails on CUB (mIoU: 10.7) and Flowers (mIoU: 5.1) datasets, since fine-grained classes do not contain diverse objects. 

\begin{table}[tb]
\begin{center}
\caption{Different auxiliary datasets to generate pseudo labels. Mean IoU is reported for both tasks on 15-5 scenario.}
\resizebox{0.45\textwidth}{!}{%
\begin{tabular}{c||c|cc|c}
                           &               & \multicolumn{3}{c}{\textbf{15-5}} \\
Candidate Dataset & FID Score  & 1-15   & 16-20   & all             \\ \hline
COCO                       & \textbf{25.8} & \textbf{76.9}   & \textbf{54.3}    & \textbf{71.3}   \\
ImageNet (Val)              & 43.8          & 75.5   & 53.5    & 70.0            \\
CUB                        & 153.5         & 4.0    & 30.9    & 10.7            \\
Flowers                    & 222.3         & 2.6    & 12.8    & 5.1\\       
\end{tabular}}
\label{tab:fid}
\end{center}
\end{table}

\tableade  

\figade 

We did a preliminary experiment on measuring the relatedness of datasets mathematically with Fréchet Inception Distance (FID) score\cite{heusel2017gans}. FID is originally proposed to measure the similarity of two distributions, such as generated fake images and real images, which is popular in measuring the quality of generative models.  We use FID here to measure the closeness between labeled datasets and unlabeled datasets. We compute FID between Pascal-VOC and other unlabeled datasets, we obtain an FID score with COCO (25.8), ImageNet (43.8), CUB (153.5) and Flowers (222.3). Smaller FID scores indicate that the distribution of two datasets are closer, which is consistent with performance we obtained for self-training. Therefore such a measure could be used to select good auxiliary data. 

\subsection{On ADE20K}

Following~\cite{cermelli2020modeling}, we report average mIoU of two different class orders on ADE20K dataset. In this experiment, we only compare with activation-based regularization methods LwF, LwF-MC, ILT (much better than PI, EWC and RW), the state-of-the-art method MiB and MiB with auxiliary data (MiB + Aux). Here we consider three scenarios: 100-50 means 100 classes for the first task and 50 classes for the second tasks. 100-10 considers 100 classes as the first task and the rest classes are divided into 5 tasks. Lastly, in 50-50 scenario, 150 classes are equally distributed in three tasks with 50 classes each.

\subsubsection{Single-step addition of 50 classes (100-50).} As shown in Table~\ref{tab:ade}, FT forgets the first task totally because of the large number of classes. As seen from Joint, the overall mIoU is 38.9\%, which is much less compared to Pascal-VOC 2012, indicating this is a more challenging dataset. MiB achieves relatively robust results in this scenario, while our method outperforms it by 0.5\% in average for 150 classes. Notably, our method prevents the forgetting of previous task effectively but obtains worse performance on the second task compared to MiB. Interestingly, we have seen that for Pascal-VOC dataset, MiB + Aux outperforms baseline MiB in most cases, while it fails to further improve the performance in this case. The reason could be that ADE-20K is more challenging and accuracy itself is much lower than on Pascal-VOC, which could introduce more noise during training.
It further shows the importance of our proposed framework to leverage unlabeled data, which leads to superior performance. 

We report some qualitative results for different incremental methods (Ours, MiB and Fine-tuning) on this senario in Fig.~\ref{fig:ade}. On this challenging scenario with more categories, our approach is capable of segmenting more objects correctly (e.g. the wall, apparel, floor, painting, box, computer) than MiB.  

\subsubsection{Multi-step of addition 50 classes (100-10).} This is a more challenging scenario with six tasks in total. Most methods fail in this scenario, only achieving about 1.0\% mIoU (FT, LwF and ILT). ILT performs better than LwF-MC in most scenarios on Pascal-VOC but worse results on ADE20K. Our method still outperforms MiB overall by 2.2\%. More specifically, the gain is significant for the first four tasks from 1.8\% (first task) to 11.3\% (third task). And the performance of the fifth task is comparable, while ours obtains worse result for the last task. Again, using auxiliary data for MiB (MiB + Aux) leads to worse results compared to the original MiB, which shows that without specific design, the unlabeled data can be harmful.

\subsubsection{Three steps of 50 classes (50-50).} This is a more balanced scenario where each training session has the same number of classes. Similar to previous scenarios, we improve overall mIoU of MiB from 27.0\% to 29.0\%. Specifically, the gain is 4.5\% for the first task and 3.9\% for the second task. One general observation for all scenarios is that there is still a large gap between incremental learning and Joint for the ADE dataset. This suggests that incremental segmentation learning has to be further developed and improved.  

\section{Conclusions}

In this work, we improve incremental semantic segmentation with self-training. Unlabeled data is leveraged to combine the knowledge of the previous and current models. Importantly, conflict reduction provides more accurate pseudo labels for self-training. We have achieved state-of-the-art performance on two large datasets Pascal-VOC 2012 and ADE20K. We show that our method can obtain superior results on Pascal-VOC 2012 using only 1\% unlabeled data. Qualitative results show significant  more accurate segmentation maps are generated compared to the other methods. 

There are still several limitations of our method. As seen from the experimental results, our proposed method works relatively better on VOC than ADE, one reason is due to the task difficulty of different datasets. Our method benefits more from a better segmentation model itself, which can provide better pseudo labels. We have also shown that our method can learn from general unlabeled datasets, such as ImageNet, however, large domain differences can still be a challenging problem for using unlabeled data. 

As possible future directions, involving human monitors in the loop of using unlabeled data has potential to reduce the errors in real applications. Moreover, it would be interesting to explore other ways of replay to avoid catastrophic forgetting, such as storing raw exemplars from previous data, or generating object templates for replay. Besides, the development of the field of pseudo labelling is complementary to our method, which can further improve the performance of the incremental learning setting for semantic segmentation.

\section*{Acknowledgment}
We acknowledge the support from Huawei Kirin Solution, the Spanish Government funding for projects
PID2019-104174GB-I00.

\ifCLASSOPTIONcaptionsoff
  \newpage
\fi

\bibliographystyle{IEEEtran}
\bibliography{IEEEabrv, tnnls}

\begin{IEEEbiography}[{\includegraphics[width=1in,height=1.25in,clip,keepaspectratio]{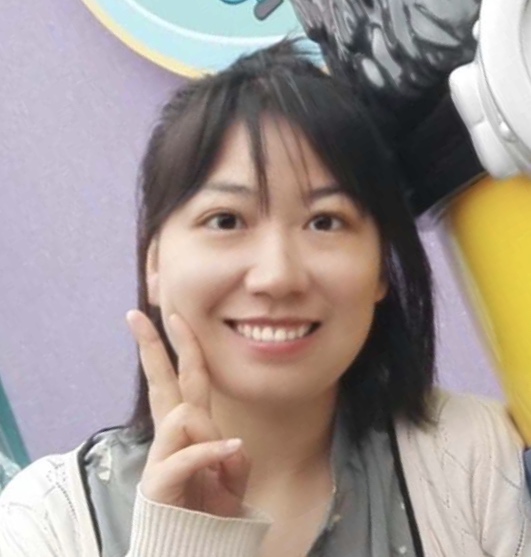}}]{Lu Yu} is currently an associate professor at Tianjin University of Technology, Tianjin, China. Before that She was a pos-doc at Heriot Watt University, Edinburgh, UK. She received her Ph.D in computer science from Autonomous University of Barcelona, Barcelona, Spain in 2019 and master degree from Northwestern Polytechnical University in 2015, Xi'an, China.  Her research interests include lifelong learning, metric learning, multi-model learning and color representation learning.
\end{IEEEbiography}

\begin{IEEEbiography}[{\includegraphics[width=1in,height=1.25in,clip,keepaspectratio]{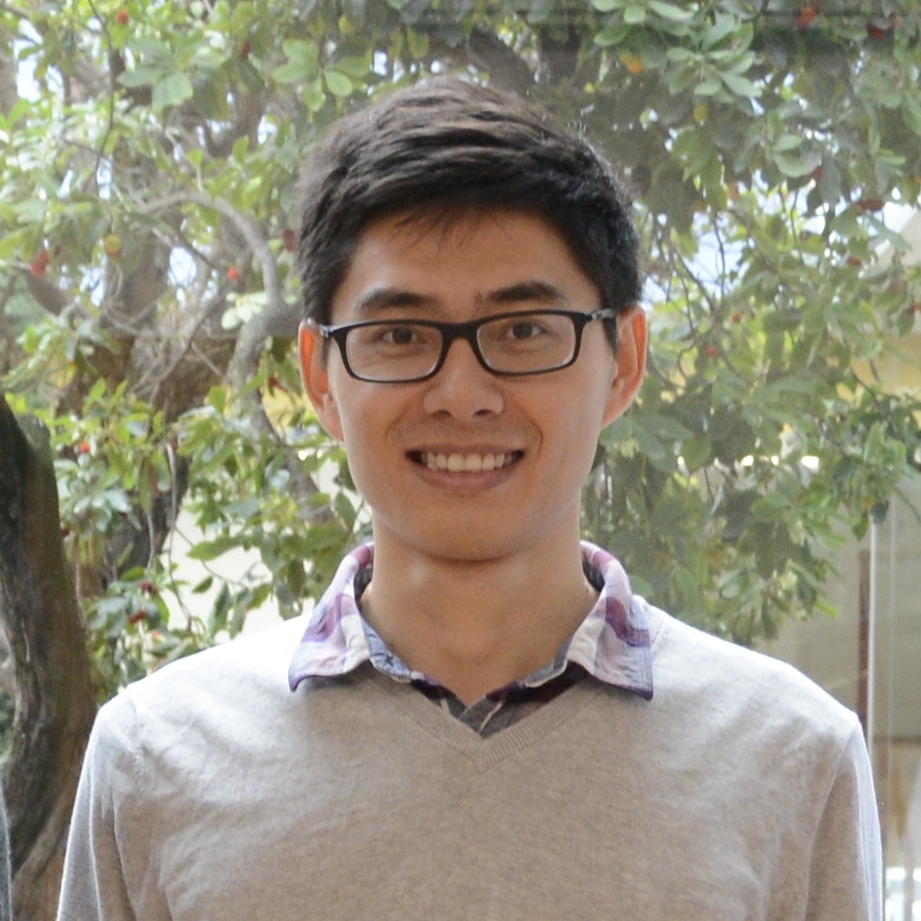}}]{Xialei Liu} is currently an associate professor at Nankai University, Tianjin, China. Before that, he was a post-doc research associate at University of Edinburgh, Edinburgh, UK. He obtained his PhD at the Autonomous University of Barcelona in 2020, Barcelona, Spain. He received B.S. and M.S. degrees at Northwestern Polytechnical University in 2013 and 2016, respectively, Xi'an, China. His research interests include lifelong learning, self-supervised learning, few-shot learning, long-tailed learning, and many applications (classification, detection, segmentation, crowd counting, image quality assessment, etc).
\end{IEEEbiography}

\begin{IEEEbiography}[{\includegraphics[width=1in,height=1.25in,clip,keepaspectratio]{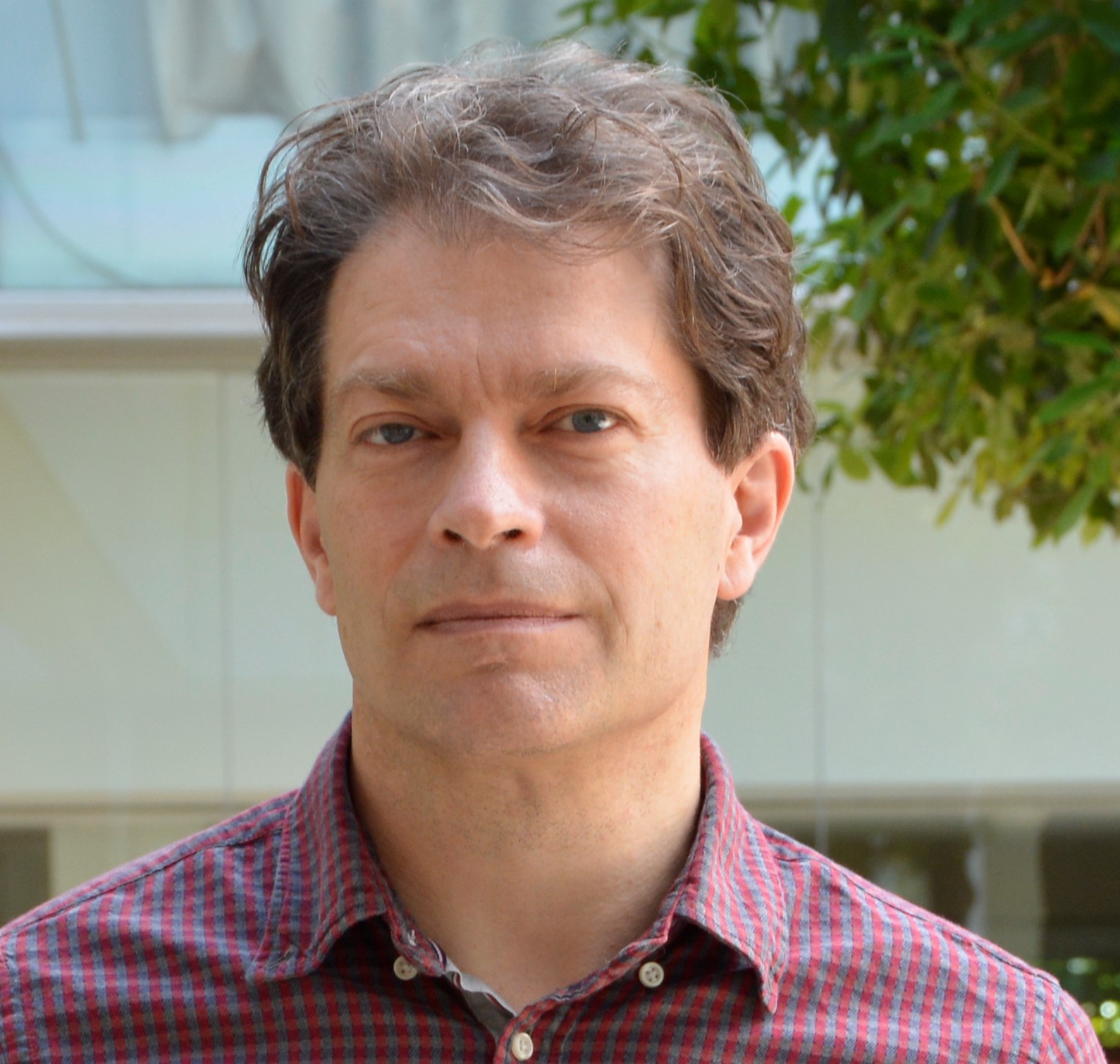}}]{Joost van de Weijer} received the Ph.D. degree from the University of Amsterdam in 2005, Amsterdam, Netherlands. He was a Marie Curie Intra-European Fellow at INRIA Rhone-Alpes, France, and from 2008 to 2012 was a Ramon y Cajal Fellow at the Universitat Autònoma de Barcelona, Barcelona, Spain, where he is currently a Senior Scientist at the Computer Vision Center and leader of the Learning and Machine Perception (LAMP) Team. His main research directions are color in computer vision, continual learning, active learning, and domain adaptation.
\end{IEEEbiography}

\end{document}

%% file: resources.tex
\newcommand{\tablepascal}{%
\begin{table*}[t]
\centering
\setlength{\tabcolsep}{4pt} 
\small
\caption{Mean IoU on the Pascal-VOC 2012 dataset for different incremental class learning scenarios. MiB is the state-of-the-art method, and MiB + Aux is MiB further trained with unlabeled data by generating pseudo labels.}
\label{tab:pascal}
\begin{tabular}{l||cc|c||cc|c||cc|c||cc|c||cc|c||cc|c}
\multicolumn{1}{c}{}    & \multicolumn{6}{c}{\textbf{{19-1}}}   & \multicolumn{6}{c}{{\textbf{15-5}}} & \multicolumn{6}{c}{{\textbf{15-1}}}    \\
\multicolumn{1}{c||}{}    & \multicolumn{3}{c||}{\bf{Disjoint}}        & \multicolumn{3}{c||}{\bf{Overlapped}}  & \multicolumn{3}{c||}{\bf{Disjoint}}     & \multicolumn{3}{c||}{\bf{Overlapped}}  & \multicolumn{3}{c||}{\textbf{Disjoint}}      & \multicolumn{3}{c}{\textbf{Overlapped}} \\
\bf{Method} & \it{1-19}  & \it{20}   & \it{all}  & \it{1-19}  & \it{20}   & \it{all}     & \it{1-15}  & \it{16-20}   & \it{all}   & \it{1-15}  & \it{16-20}   & \it{all} & \it{1-15}  & \it{16-20}   & \it{all}  & \it{1-15}  & \it{16-20}   & \it{all}     \\ \hline
{FT }     & 5.8     & 12.3    & 6.2     & 6.8       & 12.9      & 7.1      & 1.1    & 33.6    & 9.2      & 2.1     & 33.1  & 9.8    & 0.2        & 1.8        & 0.6        & 0.2        & 1.8        & 0.6       \\
{PI}     & 5.4     & 14.1    & 5.9     & 7.5       & 14.0      & 7.8      & 1.3    & 34.1    & 9.5      & 1.6     & 33.3  & 9.5    & 0.0 &	1.8 &	0.4   & 0.0        & 1.8        & 0.5 \\
{EWC}    & 23.2    & 16.0    & 22.9    & 26.9      & 14.0      & 26.3     & 26.7   & 37.7    & 29.4     & 24.3    & 35.5  & 27.1   & 0.3        & 4.3        & 1.3        & 0.3        & 4.3        & 1.3       \\
{RW}     & 19.4    & 15.7    & 19.2    & 23.3      & 14.2      & 22.9     & 17.9   & 36.9    & 22.7     & 16.6    & 34.9  & 21.2   & 0.2        & 5.4        & 1.5          & 0.0        & 5.2        & 1.3 \\
{LwF}    & 53.0    & 9.1     & 50.8    & 51.2      & 8.5       & 49.1     & 58.4   & 37.4    & 53.1     & 58.9    & 36.6  & 53.3   & 0.8        & 3.6        & 1.5        & 1.0        & 3.9        & 1.8       \\
{LwF-MC} & 63.0    & 13.2    & 60.5    & 64.4      & 13.3      & 61.9     & 67.2   & 41.2    & 60.7     & 58.1    & 35.0  & 52.3   & 4.5        & 7.0        & 5.2    & 6.4        & 8.4        & 6.9 \\
{ILT}    & 69.1    & 16.4    & 66.4    & 67.1      & 12.3      & 64.4     & 63.2   & 39.5    & 57.3     & 66.3    & 40.6  & 59.9   & 3.7        & 5.7        & 4.2        & 4.9        & 7.8        & 5.7       \\
MiB   & 69.6 & 25.6   & 67.4      & 70.2    & 22.1       & 67.8    & 71.8     & 43.3    & 64.7    & 75.5     & 49.4  & 69.0  & 46.2       & 12.9       & 37.9       & 35.1       & 13.5       & 29.7 \\ 

MiB + Aux  & 68.6  & 28.0     & 66.6       &  70.2   &  33.4    & 68.3  &   73.1   &  49.6  & 67.2   &  74.2  & 52.9  & 68.9 &    48.0   &    15.7   & 39.9 & 42.5 & 21.8 & 37.3             \\ 

Ours    & \bf{76.6} & \bf{36.0}   & \bf{75.4}      & \bf{76.1}    & \bf{43.4}       & \bf{74.5}    & \bf{76.9}     & \bf{54.3}    & \bf{71.3}    & \bf{76.7}     & \bf{54.3}  & \bf{71.1}  & \bf{70.1}       & \bf{34.3}       & \bf{61.2}       & \bf{71.4}       & \bf{40.0}       & \bf{63.6}             \\ 
\hline
{Joint} & 77.4&	78.0&	77.4&	77.4&	78.0&	77.4&	79.1&	72.6&	77.4&	79.1&	72.6&	77.4 & 79.1       & 72.6       & 77.4       & 79.1       & 72.6       & 77.4
\end{tabular}
\end{table*}
}

\newcommand{\tableade}{%
\begin{table*}[t]
\centering
\small
\setlength{\tabcolsep}{4pt} 
\caption{Mean IoU on the ADE20K dataset for different incremental class learning scenarios.}
\label{tab:ade}
\begin{tabular}{l||cc|c||cccccc|c||ccc|c}
\multicolumn{1}{c}{} & \multicolumn{3}{c}{{\textbf{100-50}}} & \multicolumn{7}{c}{{\textbf{100-10}}} & \multicolumn{4}{c}{{\textbf{50-50}}} \\
\textbf{Method}       & \textit{1-100} & \textit{101-150} & \textit{all}  & \textit{1-100} & \textit{100-110} & \textit{110-120} & \textit{120-130} & \textit{130-140} & \textit{140-150} & \textit{all}  & \textit{1-50} & \textit{51-100} & \textit{101-150} & \textit{all}  \\ \hline
FT     & 0.0   & 24.9    & 8.3  & 0.0   & 0.0    & 0.0    & 0.0    & 0.0    & 16.6   & 1.1  & 0.0  & 0.0    & 22.0    & 7.3  \\
LwF    & 21.1  & 25.6    & 22.6 & 0.1   & 0.0    & 0.4    & 2.6    & 4.6    & 16.9   & 1.7  & 5.7  & 12.9   & 22.8    & 13.9 \\
LwF-MC  & 34.2  & 10.5    & 26.3 & 18.7  &	2.5 &	8.7 &	4.1 &	6.5 &	5.1 &	14.3  & 27.8 & 7.0    & 10.4    & 15.1 \\
ILT  & 22.9  & 18.9    & 21.6 & 0.3   & 0.0    & 1.0    & 2.1    & 4.6    & 10.7   & 1.4  & 8.4  & 9.7    & 14.3    & 10.8 \\
MiB   & 37.9  & \textbf{27.9}    & 34.6 & 31.8  & 10.4 & 14.8   & 12.8   & 13.6   & \textbf{18.7}   & 25.9 & 35.5 & 22.2  & \textbf{23.6}    & 27.0 \\ 
MiB + Aux   &  29.0 & 28.0   & 28.7  & 24.0 & 8.3 & 7.4   & 6.2   & 4.0   & 11.9   & 18.9 & 20.2
& 21.5
  & 23.5
   & 21.8   \\ 
Ours   & \textbf{40.7}  & 24.0    & \textbf{35.1} & \textbf{33.6}  & \textbf{18.7}   & \textbf{25.5}   & \textbf{16.7}   & \textbf{13.7}   & 9.7   & \textbf{28.1} & \textbf{40.0} & \textbf{26.1}   & 21.0    & \textbf{29.0} \\
\hline
Joint  & 44.3  & 28.2    & 38.9 & 44.3  & 26.1   & 42.8   & 26.7   & 28.1   & 17.3   & 38.9 & 51.1 & 38.3   & 28.2    & 38.9
\end{tabular}
\end{table*}
}

\newcommand{\tableablation}{%
\begin{table*}[t]
\small
\centering
\setlength{\tabcolsep}{3pt} 
\caption{Ablation study of our method on Pascal-VOC 2012 \textit{disjoint} setup. \textit{ST*}, \textit{ST}, \textit{CR} and \textit{MS} denote self-training variant, our self-training, conflict reduction and maximizing self-entropy, respectively.} 
\label{tab:ablation}
\resizebox{0.6\textwidth}{!}{%
\begin{tabular}{l||cc|c||cc|c||cc|c}
      \multicolumn{1}{c}{}& \multicolumn{3}{c}{\textbf{19-1}}          & \multicolumn{3}{c}{\textbf{15-5}}          & \multicolumn{3}{c}{\textbf{15-1}} \\
  &\it{1-19} & \it{20} & \it{all} & \it{1-15} & \it{16-20} & \it{all} & \it{1-15}  & \it{16-20}     & \it{all}     \\ \hline
FT          & 5.8      & 12.3       & 6.2       & 1.1      & 33.6       & 9.2       & 0.2         & 1.8   & 0.6   \\
MiB          & 69.6     & 25.6       & 67.4       & 71.8      & 43.3       & 64.7       & 46.2         & 12.9   & 37.9   \\ \hline
 \textit{ST*} & 72.5  &  20.0   & 69.9  &  55.0 &   37.1 &   50.5 & 26.9    &  12.7 &  23.3 \\
     \textit{ST} & 75.9  & 35.8    &  74.8 & 78.0 & 30.9       &     66.3   & 67.9         &  18.6  & 55.6   \\
\textit{ST}+\textit{CR}&  76.2   &    34.4   &    75.0   &   76.0  & 53.9  & 70.5 & 69.0 & 28.1  & 58.8 \\
\textit{ST}+\textit{CR}, \textit{MS} & \textbf{76.6}      & \textbf{36.0}      & \textbf{75.4}      & \textbf{76.9}      & \textbf{54.3}      & \textbf{71.3}      & \textbf{70.1} & \textbf{34.3} & \textbf{61.2} \\
\hline
Joint         & 77.4      & 78.0       & 77.4   &  79.1      & 72.6       & 77.4       & 79.1         & 72.6   & 77.4   \\
\end{tabular}}
\end{table*}
}

\newcommand{\tablealgorithm}{%
\begin{algorithm}[tb]
\caption{: Self-training for semantic segmentation CIL. }
\label{table:algorithm}
\begin{algorithmic}[1]
\STATE\textbf{Input:} Sequence $\mathbfcal{T}^1, \ldots ,  \mathbfcal{T}^t, \ldots , \mathbfcal{T}^T$,   where $\mathbfcal{T}^t = \left(\mathbf{X}^t,\mathbf{Y}^t\right)$. And auxiliary data $\mathbf{A}$. \\
\STATE\textbf{Output:} A model $f_{\theta^T}$ after learning $T$ tasks incrementally. \\

\FOR{$t=1, \ldots , T$} 
\IF{$t = 1$} 
\STATE Train a standard semantic segmentation model $f_{\theta^1}$  using Eq.~\ref{eq:obj-general} or Eq.~\ref{eq:total} if self-entropy maximization is applied. 
\ELSE
\STATE Train a temporary model $f_{\theta^*}$ (initialized from the old model $f_{\theta^{t-1}}$) using current data $\mathbf{X}^{t}$.

\STATE Given $f_{\theta^{t-1}}$ and $f_{\theta^*}$, we introduce auxiliary data $\mathbf{A}$. Pseudo labels are predicted by both models represented as   $\hat{P}^{t-1}$ and $\hat{P}^{t}$ respectively. 

\STATE Final pseudo labels $\hat{P}^{t}_{\mathbf{A}}$ are fused by $\hat{P}^{t-1}$ and $\hat{P}^{t}$ using the proposed conflict reduction mechanism (Eq.~\ref{eq:conflict}). 

\STATE A new model $f_{\theta^{t}}$ (initialized from the old model $f_{\theta^{t-1}}$) is updated using auxiliary data $A$ and pseudo labels $\hat{P}^{t}_{\mathbf{A}}$.
\ENDIF

\ENDFOR
\end{algorithmic}
\end{algorithm}
}

\newcommand{\figvoc}{%
\begin{figure*}[tb]
\begin{center}
\includegraphics[width=1\textwidth]{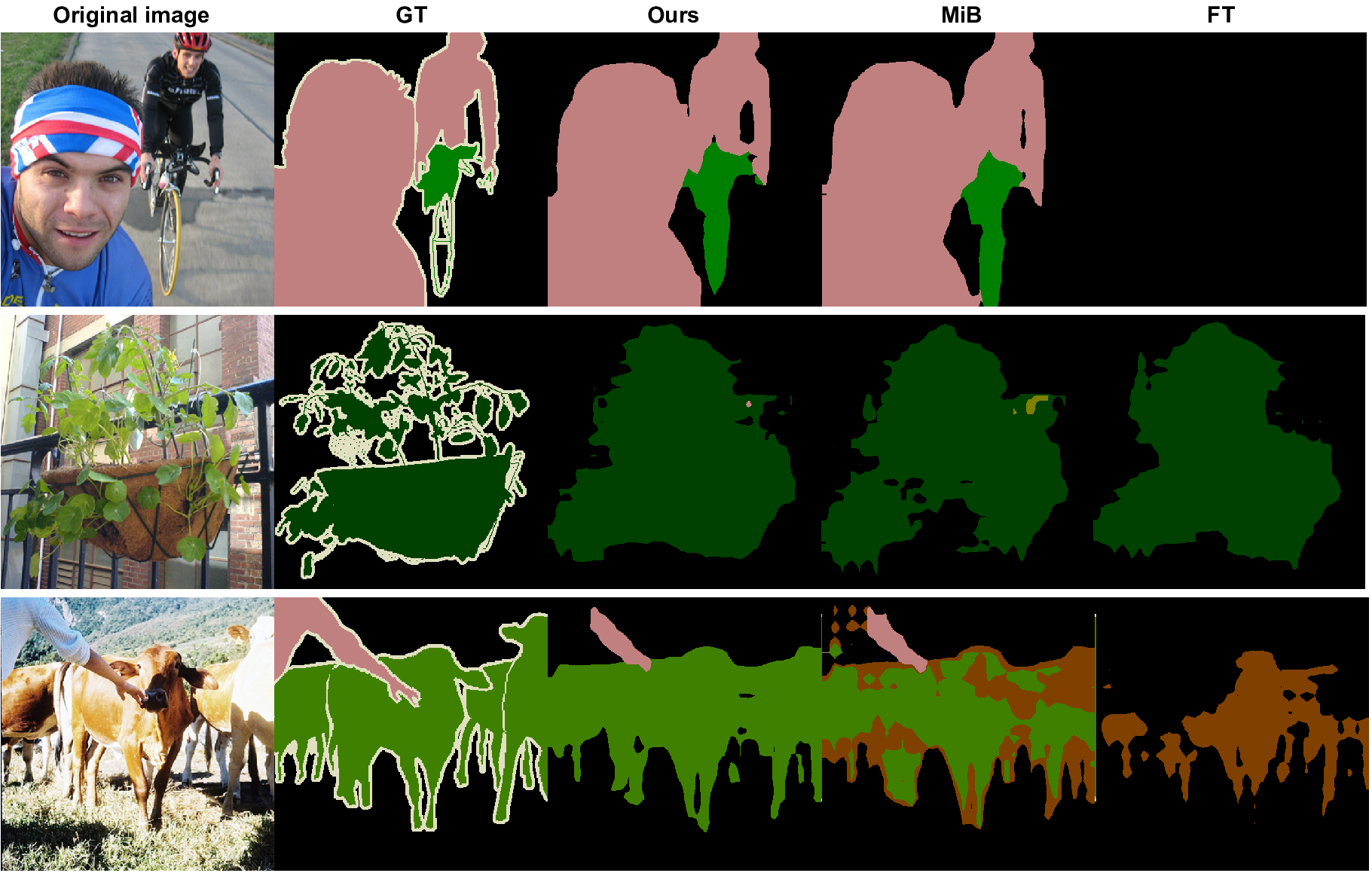}
\caption{Visualization on Pascal-VOC 2012 with 15-5 scenario. From left to right: original image, ground-truth, Ours, MiB and FT. 
We can see that our method has higher quality predictions compared to  state-of-the-art CIL method MiB.
}
\label{fig:voc}
\end{center}
\end{figure*}
}

\newcommand{\figade}{%
\begin{figure*}[tb]
\begin{center}
\includegraphics[width=0.99\textwidth]{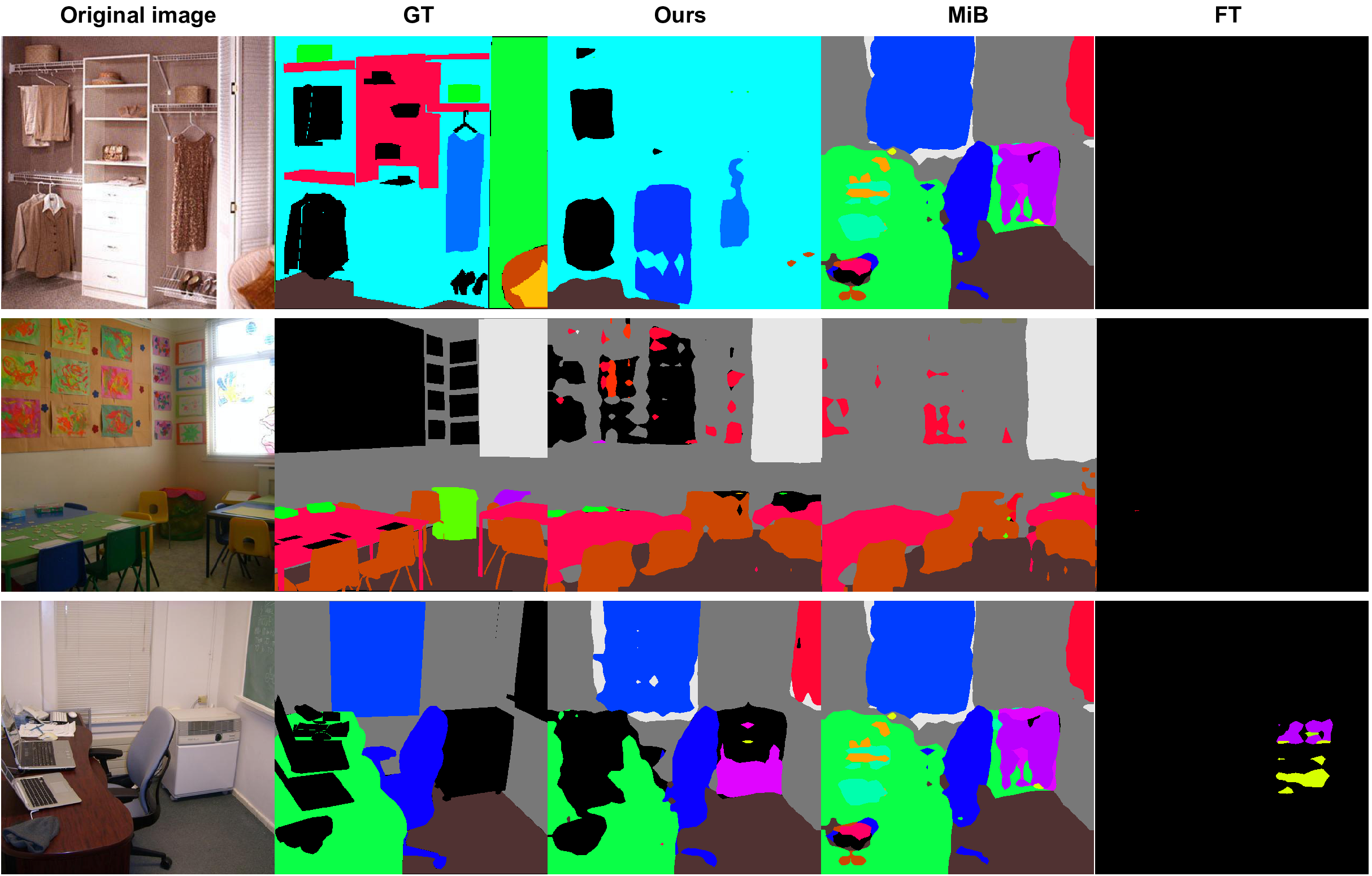}
\caption{Visualization on ADE20K with 100-50 scenario. From left to right: original image, ground-truth, Ours, MiB and FT. It is clear that Ours obtains the most similar results to the ground truth.
}
\label{fig:ade}
\end{center}
\end{figure*}
}

\newcommand{\figintro}{%
\begin{figure}[tb]
\begin{center}
\includegraphics[width=0.49\textwidth]{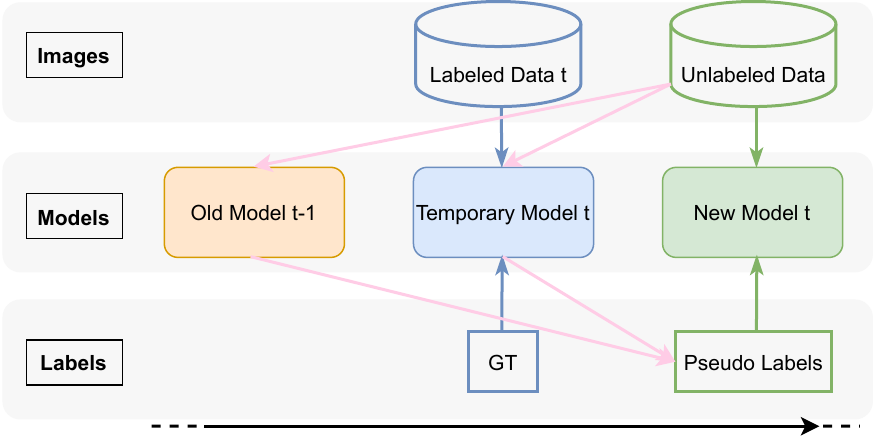}
\caption{\xl{Illustration of our method for class-incremental semantic segmentation. At training session $t$ in the incremental learning setting, the old model is copied from the previous task $t-1$ and a temporary model is first learned with labeled data and their ground truth (GT) available from the current task $t$ (note that the data from previous session $t-1$ is unavailable at session $t$). Then a new model is trained with unlabeled data and their pseudo labels, where pseudo labels are fused from the old model and the temporary model (denoted with pink arrows) proposed in Section~\ref{sec:method}. }}
\label{fig:intro}
\end{center}
\end{figure}
}

\newcommand{\figconflict}{%
\begin{figure}[tb]
\begin{center}
\includegraphics[width=0.49\textwidth]{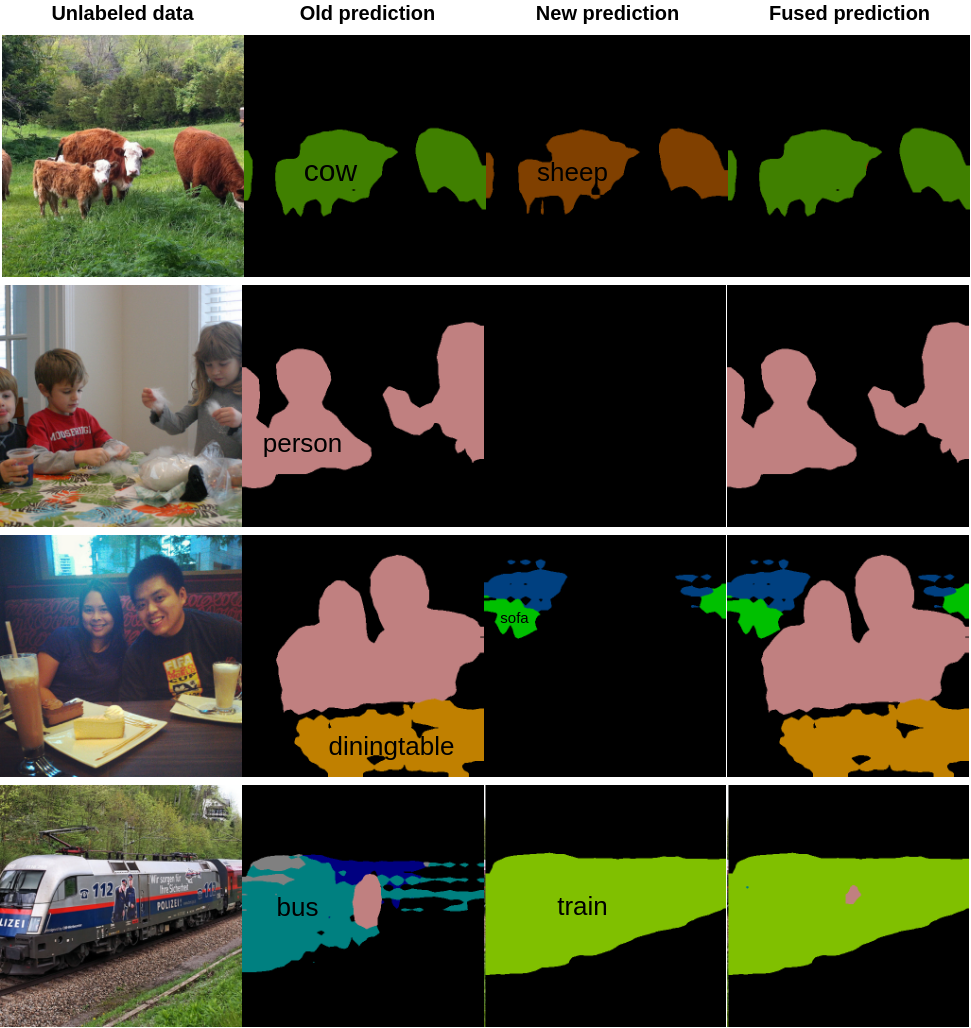}
\caption{Illustration of challenges while fusing predictions from both old and new models for incremental semantic segmentation. The data (first column) and its predictions by both the old (second column) and temporary model (third column). Accurate pseudo label fusion is achieved by the proposed conflict reduction mechanism (last column). Colors indicate different categories and background is denoted in black. }
\label{fig:v-conflict}
\end{center}
\end{figure}
}